\documentclass{article}

\usepackage{PRIMEarxiv}

\usepackage[utf8]{inputenc} % allow utf-8 input
\usepackage[T1]{fontenc}    % use 8-bit T1 fonts
\usepackage{hyperref}       % hyperlinks
\usepackage{url}            % simple URL typesetting
\usepackage{booktabs}       % professional-quality tables
\usepackage{amsfonts}       % blackboard math symbols
\usepackage{nicefrac}       % compact symbols for 1/2, etc.
\usepackage{microtype}      % microtypography
\usepackage{lipsum}
\usepackage{fancyhdr}       % header
\usepackage{graphicx}       % graphics
\graphicspath{{media/}}     % organize your images and other figures under media/ folder

\usepackage{listings}
\usepackage{makecell}
\usepackage{tabularx}
\usepackage{adjustbox}

\title{JORLDY: A Fully Customizable Open Source Framework for Reinforcement Learning
}

\author{
  Kyushik Min, Hyunho Lee, Kwansu Shin, Taehak Lee, Hojoon Lee, Jinwon Choi, Sungho Son \\
  Kakao Enterprise \\
  Seongnam, South Korea\\
  \texttt{\{leonard.q, ramanuzan.lee, kan.s, erinn.lee, link.lee, royce.choi, crest.son\}} \\ 
  \texttt{@kakaoenterprise.com} \\
}

\begin{document}
\maketitle

\begin{abstract}
Recently, Reinforcement Learning (RL) has been actively researched in both academic and industrial fields. However, there exist only a few RL frameworks which are developed for researchers or students who want to study RL. In response, we propose an open-source RL framework "Join Our Reinforcement Learning framework for Developing Yours" (JORLDY). JORLDY provides more than 20 widely used RL algorithms which are implemented with Pytorch. Also, JORLDY supports multiple RL environments which include OpenAI gym, Unity ML-Agents, Mujoco, Super Mario Bros and Procgen. Moreover, the algorithmic components such as agent, network, environment can be freely customized, so that the users can easily modify and append algorithmic components. We expect that JORLDY will support various RL research and contribute further advance the field of RL. The source code of JORLDY is provided on the following Github: \url{https://github.com/kakaoenterprise/JORLDY} 
\end{abstract}

% keywords can be removed

\section{Introduction}
Reinforcement Learning (RL) has shown impressive performance in various fields including game \cite{silver2017mastering,vinyals2019grandmaster,badia2020agent57}, chip placement \cite{mirhoseini2021graph}, energy \cite{deepmind2016reduces}, recommendation \cite{afsar2021reinforcement}, robotics \cite{finn2016deep,nair2018visual} and many others. Therefore, RL has been actively researched on both academic and industrial fields lately. 
Following this trend, several RL frameworks have been released recently. For the researchers and students who study RL, the algorithms and environments of RL framework should be customizable to develop their research. For the user, who is not a RL expert and wants to use RL framework for industrial problem, the RL framework should be easy to use and be able to add new industrial environments. Accordingly, there has been a growing demand for a RL framework that can be easily used and customized.

There are some RL frameworks \cite{baselines, plappert2016kerasrl, liang2017ray, hoffman2020acme, weng2021tianshou, fujita2021chainerrl} opened for the users who wants to study or use RL algorithms. However, Some frameworks have problems that hinder usability or convenience for the users. First, it is difficult to modify or add RL algorithms or environments in most of the frameworks. If there are restrictions on the modification of the environment or algorithm, it is difficult to conduct more advanced research based on the framework. Second, some of the frameworks such as \cite{baselines}, \cite{plappert2016kerasrl} provide less number of the algorithms than JORLDY or do not provide latest RL algorithms. Third, many existing frameworks such as \cite{baselines}, \cite{plappert2016kerasrl}, \cite{weng2021tianshou}, \cite{fujita2021chainerrl} don't provide distributed training for RL. Recently, great performance improvement can be conducted for RL algorithm with applying distributed training, so most of state-of-the-art RL algorithms have distributed training structure. Therefore, it is important for RL frameworks to support distributed training.   

For these reasons, we propose a new open source RL framework "Join Our Reinforcement learning framework for Developing Yours" (JORLDY). JORLDY is named after Jordy, one of the Niniz character of the Kakao corporation which is illustrated in Figure 1. As the full name suggests, we expect RL users to further develop their own research through the proposed framework. 

%%%%%%%%%%%%%%%%%%%%%%%%%%%%%%% Figure 1 %%%%%%%%%%%%%%%%%%%%%%%%%%%%%%%
\begin{figure}
  \centering
  \includegraphics[width=9cm]{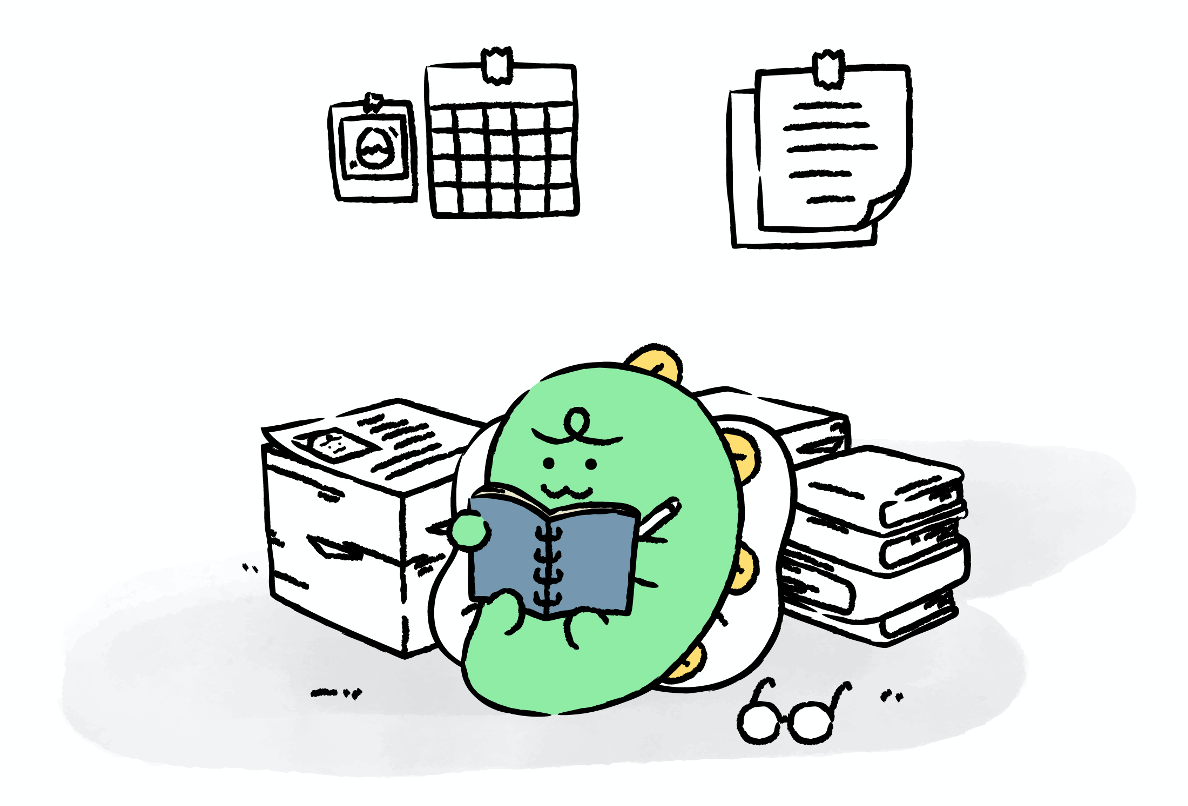}
  \caption{Niniz character Jordy of Kakao Corp.}
  \label{fig:fig1}
\end{figure}
%%%%%%%%%%%%%%%%%%%%%%%%%%%%%%%%%%%%%%%%%%%%%%%%%%%%%%%%%%%%%%%%%%%%%%%%

JORLDY framework provides more than 20 RL algorithms from basic to advanced algorithms. The algorithms consist of various kinds of RL algorithms such as value-based, policy-based, actor-critic, distributional and distributed structure algorithms. Each RL algorithm consists of some components: agent, network, buffer and optimizer. All the components of RL algorithms are fully customizable, so users can modify the existing algorithms and develop new RL algorithms based on the provided RL algorithms of JORLDY. Also, all the components can be freely combined with each other, so various implementations are possible. Moreover, JORLDY provides various distributed architecture using Ray \cite{moritz2018ray}, so it can support RL algorithms that have distributed training process. The RL algorithms are implemented using Python with deep learning library Pytorch \cite{paszke2019pytorch}. 

Also, JORLDY supports various existing RL environments such as OpenAI gym \cite{brockman2016openai}, Arcade Learning Environment (ALE) \cite{machado18arcade}, Gym Super Mario Bros \cite{gym-super-mario-bros}, Procgen \cite{cobbe2019procgen} and MuJoCo \cite{todorov2012mujoco}. For the diversity of the RL environments, some of the custom environments made with Unity ML-Agents toolkit \cite{juliani2018unity} are provided. Any environments can be applied if they set to a specified format for JORLDY. Because of this characteristic, it is expected that RL can be applied to not only games but also various industrial environments through JORLDY.

Lastly, JORLDY considers usability that non RL expert can use RL without difficulties. Therefore, the provided RL algorithms and environments can be easily connected and run with single line of command. 
% 한줄의 라인으로 실행된다는 점 외에 usability로 고려할만한 점이 있을까?? 

The main features of JORLDY are as follows: 
\begin{itemize}
\item More than 20 RL algorithms and various RL environment are provided
\item The RL components (i.e. network, agent, buffer) and environments can be added and fully customized
\item Various combination of the RL components can be simply conducted
\item All the provided algorithms and environments can be run with a simple command
\item Various distributed architecture for RL algorithms are supported using Ray  
\item Benchmark of the algorithms is conducted in many RL environment  
\end{itemize}

The rest of this paper is constructed as follows: Section 2 describes overall structure and each component of JORLDY. Section 3 describes the features of JORLDY. About usability, customizability, distributed architecture and benchmark of JORLDY is explained in this section. In section 4, we conclude this paper. 

\section{Structure of JORLDY}
% Structure의 경우 조금 더 각 component에 대한 설명이 추가되면 좋을 것 같습니다! 내용 확인 부탁드려요!

The overall structure of JORLDY is shown in Figure 2. In this section, all the components of JORLDY are described. 

%%%%%%%%%%%%%%%%%%%%%%%%%%%%%%% Figure 2 %%%%%%%%%%%%%%%%%%%%%%%%%%%%%%%
\begin{figure}
  \centering
  \includegraphics[width=0.8\textwidth]{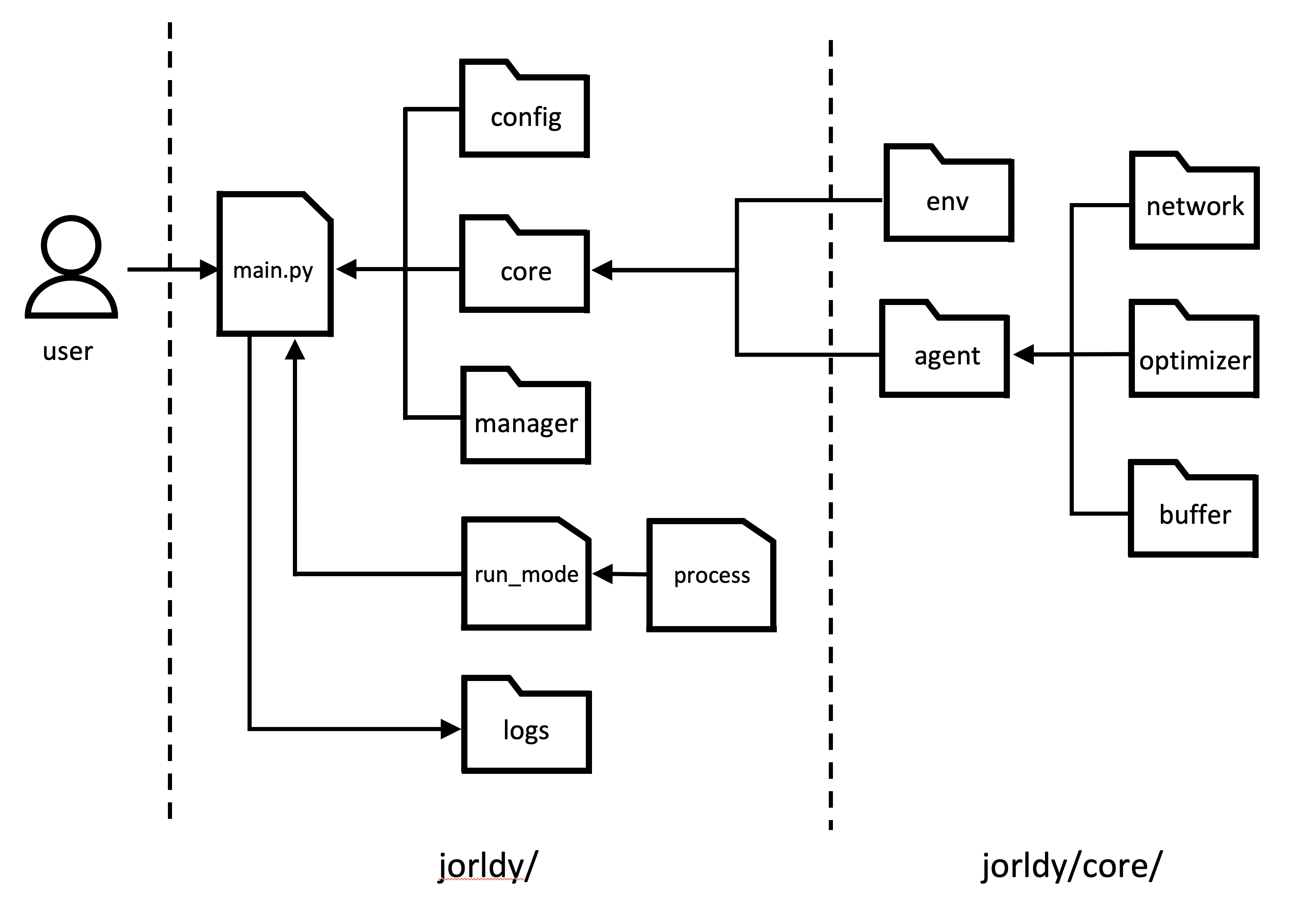}
  \caption{The entire structure of JORLDY}
  \label{fig:fig2}
\end{figure}
%%%%%%%%%%%%%%%%%%%%%%%%%%%%%%%%%%%%%%%%%%%%%%%%%%%%%%%%%%%%%%%%%%%%%%%%

At first, users have to decide train or evaluate the agent with running the script. In this process, users also can select distributed setting of the training. JORLDY has a python scripts to run: \emph{main.py}. It uses options (\emph{\texttt{-{}-}single}, \emph{\texttt{-{}-}sync}, \emph{\texttt{-{}-}async}, \emph{\texttt{-{}-}eval}) for selecting distributed training mode or evaluation. The first three options are used to train the agent. A single worker is trained without distributed setting with \emph{\texttt{-{}-}single} option. The \emph{\texttt{-{}-}sync} and \emph{\texttt{-{}-}async} options provide a different type of distributed training architecture. The \emph{\texttt{-{}-}eval} option is used to evaluate the agent. 

JORLDY has some components: \emph{config}, \emph{core}, \emph{manager}, \emph{process} and \emph{logs}. The first component of JORLDY is \emph{config}. In the \emph{config}, users can manage the parameter setting files. The configuration files are saved in the path: jorldy/config/[agent name]/[env name].py (e.g. jorldy/config/dqn/cartpole.py). The example of the configuration can be found in Appendix A. 

The second component of JORLDY is \emph{core}. The \emph{core} manages agent and environment classes. In the \emph{core}, there are 5 subcomponents: env, agent, network, optimizer and buffer. The env manages environment classes such as Gym, Atari, ML-Agents, etc. The agent manages all the agent classes such as DQN, PPO, SAC, Rainbow, etc. The network manages network classes such as noisy, policy\_value, rnd, etc. The optimizer manages optimizer classes and the buffer manages buffer class such as replay buffer, rollout buffer, per buffer, etc. All the subcomponents of \emph{core} are fully customizable. 

The other components of JORLDY are \emph{manager}, \emph{process} and \emph{logs}. The \emph{manager} manages auxiliary classes for running. The \emph{process} manages multi process functions such as interact and manage. The results of the training are saved in the jorldy/log folder. The results are saved in the path: jorldy/logs/[env name]/[agent name]/[timestamp]/ (e.g. jorldy/logs/cartpole/dqn/20211123141004). Trained model, config file, tensorboard file and gif of evaluation are saved in the path. 

\section{Features of JORLDY}

%%%%%%%%%%%%%%%%%%%%%%%%%%%%%%% Table 1 %%%%%%%%%%%%%%%%%%%%%%%%%%%%%%% 
\begin{table}[]
\centering
\caption {RL Algorithms of JORLDY}
\begin{adjustbox}{width=\textwidth}
\begin{tabular}{|c|c|}
\hline
\textbf{Category} & \textbf{Algorithms} \\ \hline
Value-based & \makecell{DQN \cite{mnih2015human}, Double DQN \cite{van2016deep}, Dueling DQN \cite{wang2016dueling}, Multistep DQN \cite{sutton2018reinforcement}, PER \cite{schaul2015prioritized},\\ Rainbow (-DQN, -IQN) \cite{hessel2018rainbow}, M-RL (DQN, IQN) \cite{vieillard2020munchausen} } \\ \hline
Policy-based, Actor-Critic & \makecell{REINFORCE \cite{sutton2000policy}, DDPG \cite{lillicrap2015continuous}, TD3 \cite{fujimoto2018addressing}, PPO \cite{schulman2017proximal}, SAC \cite{haarnoja2018soft}, MPO \cite{abdolmaleki2018maximum}, V-MPO \cite{song2019v}} \\ \hline
Distributional RL & C51 \cite{bellemare2017distributional}, QR-DQN \cite{dabney2018distributional}, IQN \cite{dabney2018implicit} \\ \hline
Exploration & Noisy DQN \cite{fortunato2017noisy}, ICM \cite{pathak2017curiosity}, RND \cite{burda2018exploration} \\ \hline
Distributed Structure & APE-X \cite{horgan2018distributed}, R2D2 \cite{kapturowski2018recurrent} \\ \hline
\end{tabular}
\end{adjustbox}
\end{table}
%%%%%%%%%%%%%%%%%%%%%%%%%%%%%%%%%%%%%%%%%%%%%%%%%%%%%%%%%%%%%%%%%%%%%%% 

\subsection{Contents of JORLDY}

JORLDY consists of various RL algorithms and environments. The implemented RL algorithms of JORLDY are shown in Table 1. According to the characteristics of algorithms, the RL algorithms are grouped as follows: value-based, policy-based \& actor-critic, distributional RL, exploration, distributed structure. The algorithms except Rainbow-IQN are implemented based on the published papers.

JORLDY supports multiple RL environments for evaluating the performance of RL algorithms. Currently, JORLDY supports following RL environments: OpenAI gym \cite{brockman2016openai}, Arcade Learning Environment (ALE) \cite{machado18arcade}, Super Mario Bros \cite{gym-super-mario-bros}, Procgen \cite{cobbe2019procgen}, Mujoco \cite{todorov2012mujoco} and Unity ML-Agents \cite{juliani2018unity}. The environments other than custom Unity ML-Agents environments are existing RL environments. 

%%%%%%%%%%%%%%%%%%%%%%%%%%%%%%% Figure 3 %%%%%%%%%%%%%%%%%%%%%%%%%%%%%%%
\begin{figure}
  \centering
  \includegraphics[width=\textwidth]{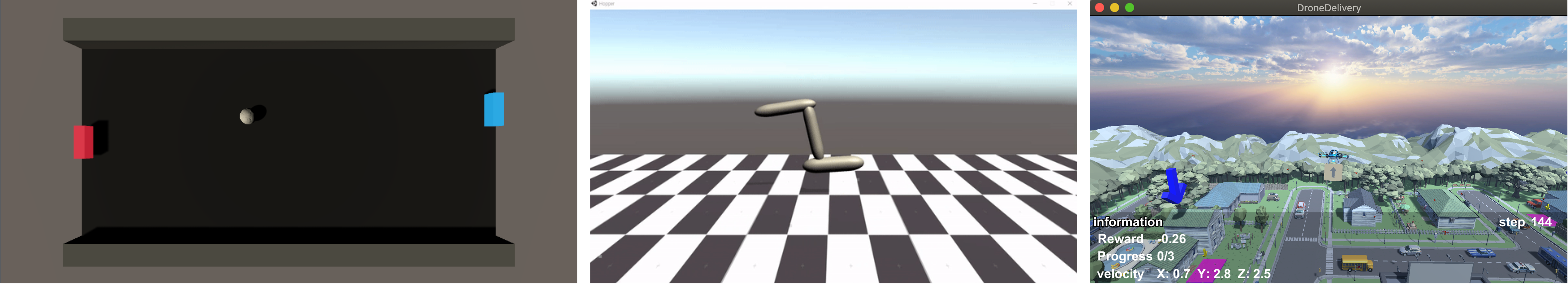}
  \caption{Examples of custom RL environments made with Unity ML-Agents (Left: Pong, Center: Hopper, Right: Drone Delivery)}
  \label{fig:fig3}
\end{figure}
%%%%%%%%%%%%%%%%%%%%%%%%%%%%%%%%%%%%%%%%%%%%%%%%%%%%%%%%%%%%%%%%%%%%%%%%

%%%%%%%%%%%%%%%%%%%%%%%%%%%%%%% Figure 4 %%%%%%%%%%%%%%%%%%%%%%%%%%%%%%%
\begin{figure}
  \centering
  \includegraphics[width=\textwidth]{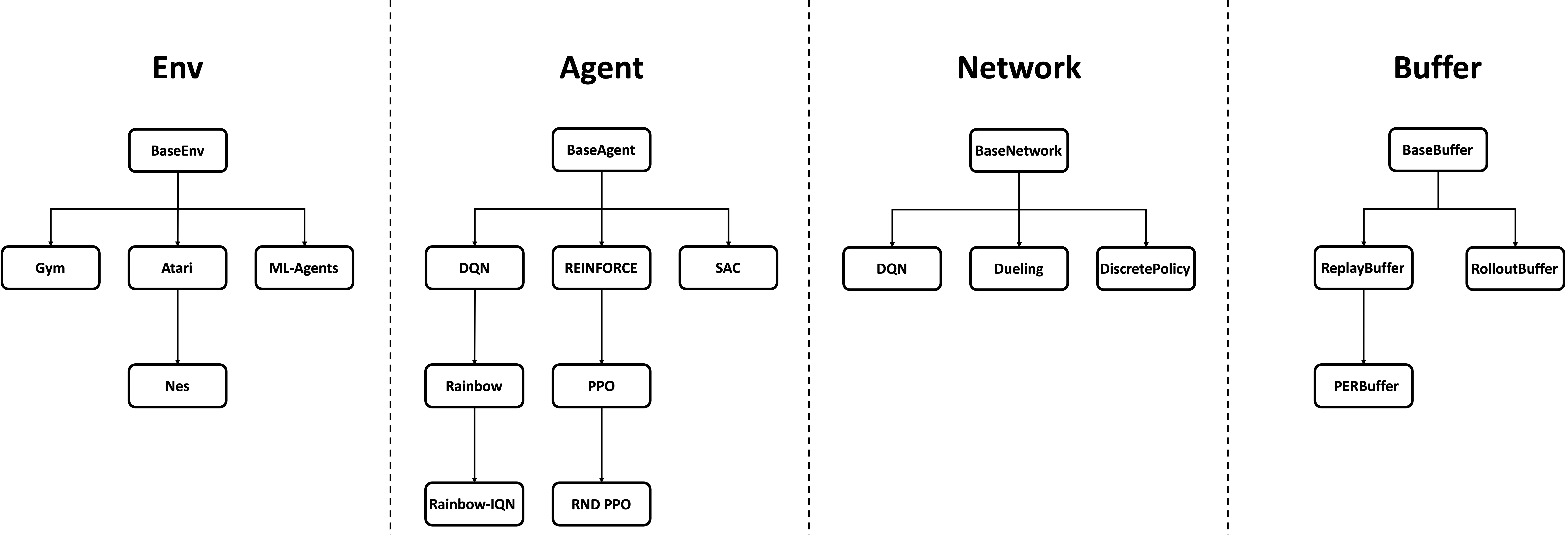}
  \caption{Code structure of JORLDY}
  \label{fig:fig4}
\end{figure}
%%%%%%%%%%%%%%%%%%%%%%%%%%%%%%%%%%%%%%%%%%%%%%%%%%%%%%%%%%%%%%%%%%%%%%%%

\subsection{Customizability}
One of the main features of JORLDY is customizability. Most of the existing RL frameworks are hard to modify the specific part of the algorithms and environments because they provide nested class structure. In the case of JORLDY, not only configuration but also all the components in the core can be modified. For example, we implemented Rainbow-IQN that uses IQN instead of C51 among the algorithms constituting Rainbow-DQN. As the case of the Rainbow-IQN, users can change few parts of existing algorithm to apply their own idea and add entire new RL algorithms. 

All the RL components such as network, buffer, agent are separately implemented, so they can be simply combined. For example, Dueling DQN, Double DQN and PER are independent algorithms in JORLDY. However, the agent of Double DQN can use network of Dueling DQN with buffer of PER with little modification. In this way, if users implement new RL component in JORLDY, the component can be applied to all the other RL components. 

Custom RL environments also can be added in JORLDY. For example, \emph{pong\_mlagent}, \emph{hopper\_mlagent}, \emph{drone\_delivery\_mlagent} of JORLDY are custom RL environment made with Unity ML-Agents. These custom environments are shown in Figure 3. Not only the environment made with Unity ML-Agents, any extisting RL environments can be applied to JORLDY.

JORLDY supports the inheritance structure for users to modify and add the algorithms and environments more conveniently. Each of the component has base class. In order to implement a new object, this base class must be inherited. However, in the network component, if necessary, an object can be implemented without inheritance of the base class (refer to icm, rnd network). In this case, the head network cannot be used. In addition, the functions of the object can be utilized by inheriting an existing object. As a result, users don't have to write unnecessary redundant code. This structure not only imposes rules on code implementation, but also makes code implementation easier. Based on the code structure, users can modify or add all the components. The example of inheritance chart is shown in Figure 4.

\subsection{Distributed Architecture}
JORLDY provides 3 different options for training: \emph{\texttt{-{}-}single}, \emph{\texttt{-{}-}sync} and \emph{\texttt{-{}-}async}. Among these options, \emph{\texttt{-{}-}single} uses a single actor for training, so it is not the one which has a distributed training structure. The \emph{\texttt{-{}-}sync} and \emph{\texttt{-{}-}async} options support different structures for distributed training. The difference between these two options is whether actors interact synchronously or asynchronously with learner. The flow chart and timeline of each distributed structure are shown in Figure 5 and Figure 6. At first, \emph{\texttt{-{}-}single} option has main process and manage process. In the main process, a single agent interacts with environments to collect transition data and trains network from it. The manage process evaluates with the latest network to get a score, and records this score and the results of training in the main process. 

The \emph{\texttt{-{}-}sync} option also has main process and manage process. In the main process, multiple actors interact in parallel at the same time to collect transition data and the learner trains the network from it. In detail, the transition data obtained through the interaction is collected and delivered to the learner when all actors complete the interact process. Then, the learner receives these data, performs training, and delivers the updated network to all actors. The manage process evaluates agent with the latest model to get a score, and records this score and the results of training in the main process. 

The \emph{\texttt{-{}-}async} option has interact process, main process and manage process. In the interact process, multiple actors interact in parallel to collect transition data. Unlike the \emph{\texttt{-{}-}sync} option, each actor interacts asynchronously. More specifically, in the \emph{\texttt{-{}-}async}, transition data is collected only for actors who have completed the interact process within a specific time and delivered to the learner. If any actor does not complete within a certain time, it waits until the completed actor comes out. Then, the learner receives these data, performs learning, and delivers the updated network only to actors who have completed the interact process. In the main process, the learner trains the model through the transition data. The manage process evaluates with the latest model to get a score, and records this score and the results of training in the main process.

%%%%%%%%%%%%%%%%%%%%%%%%%%%%%%% Figure 5 %%%%%%%%%%%%%%%%%%%%%%%%%%%%%%%
\begin{figure}
  \centering
  \includegraphics[width=13cm]{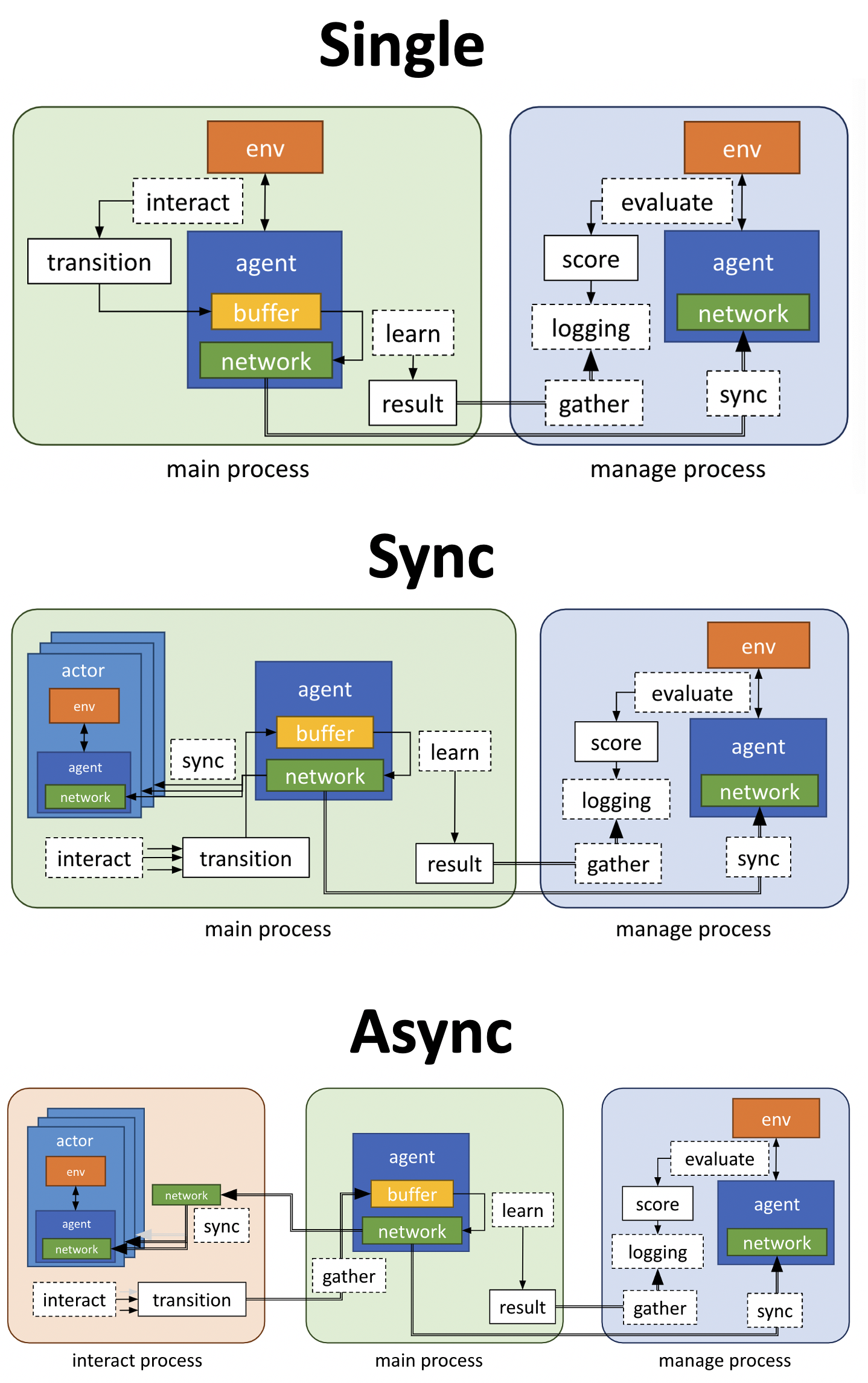}
  \caption{Flow chart of the distributed structures}
  \label{fig:fig5}
\end{figure}
%%%%%%%%%%%%%%%%%%%%%%%%%%%%%%%%%%%%%%%%%%%%%%%%%%%%%%%%%%%%%%%%%%%%%%%%

%%%%%%%%%%%%%%%%%%%%%%%%%%%%%%% Figure 6 %%%%%%%%%%%%%%%%%%%%%%%%%%%%%%%
\begin{figure}
  \centering
  \includegraphics[width=\textwidth]{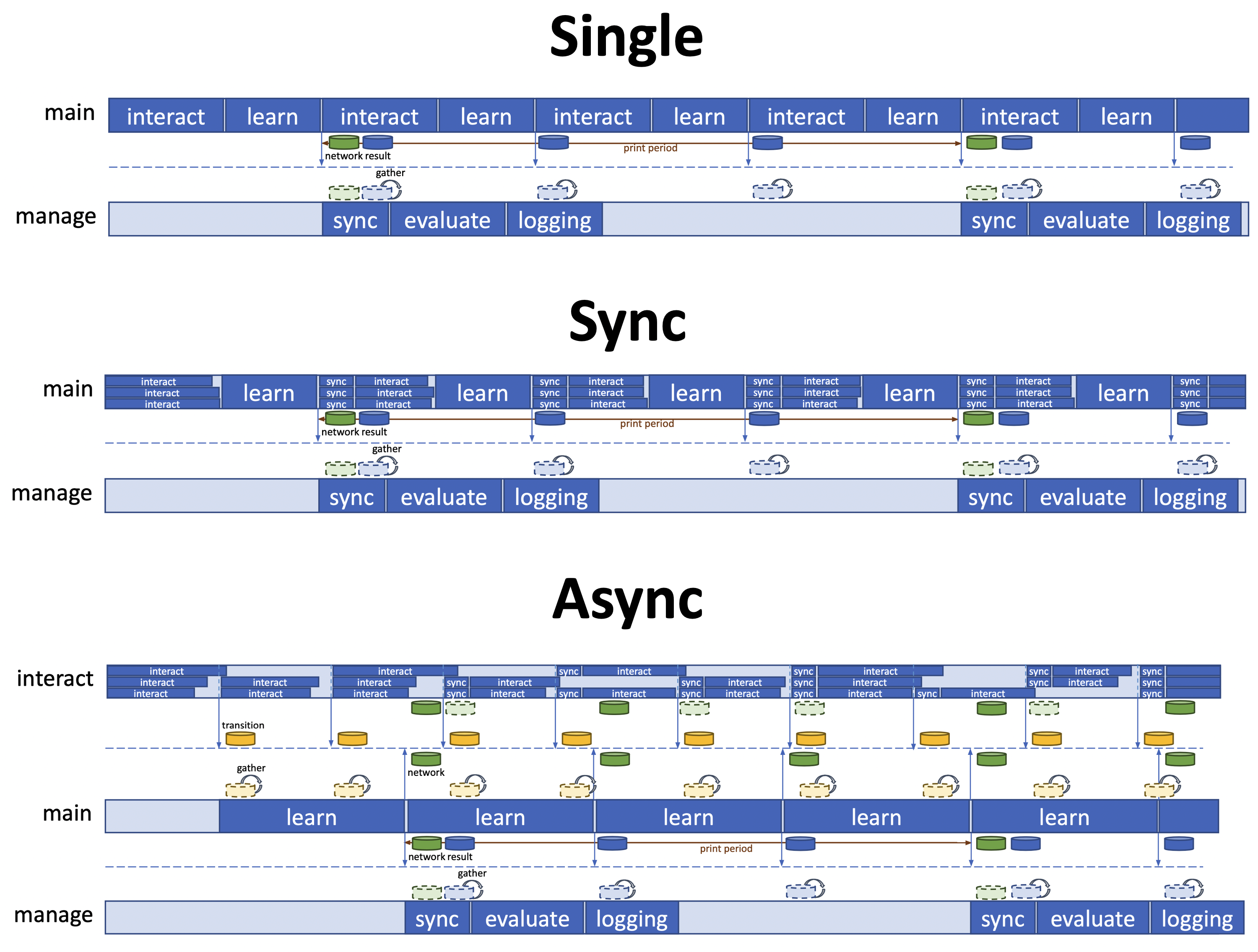}
  \caption{Timeline of the distributed structures}
  \label{fig:fig6}
\end{figure}
%%%%%%%%%%%%%%%%%%%%%%%%%%%%%%%%%%%%%%%%%%%%%%%%%%%%%%%%%%%%%%%%%%%%%%%%

\subsection{Benchmark}

Benchmark of the algorithms is conducted in RL environments. Benchmark of the Atari and MuJoCo environment can be found in Appendix B.1 and B.2, respectively. All the benchmark results of JORLDY contains not only score but also configuration, script information and specification. The results can be found in the following link: \url{https://bit.ly/3tVYYJo}  

\section{How to Use JORLDY}

Training RL algorithms in desired environment has following two process: configuration setting, executing run command. In this section, detail of the configuration setting and run command is described. 

\subsection{Configuration Setting}

At first, the user has to set the configuration of the agent and environment. As Appendix A shows, configuration files have 4 parts. In the env part, users can change the configuration of the environment. It contains information about rendering, input data (width, height, grayscale) and reward setting. In the agent part of the configuration file, hyperparameters settings of the network and agent can be adjusted. To identify the name of the network and head, the user has to check \emph{\_network\_dict.txt} and \emph{\_head\_dict.txt} in jorldy/core/network. In the optim part, user can set the optimizer and its parameters. Lastly, the configuration of training can be changed in the train part. Using the configurations, user can set whether to train or evaluate the agent. Also, various period about training, saving and printing can be changed in this part. Moreover, the distributed setting can be decided in the train part. 

\subsection{Run command}

After the configuration setting, users should run the command for executing the algorithm. Default command consists of script part and configuration part. The example command for training DQN algorithm with CartPole environment is as follows. \emph{\texttt{-{}-}single} training option is a default option, so it can be deleted. However, if user wants to train with distributed setting, \emph{\texttt{-{}-}sync} or \emph{\texttt{-{}-}async} option has to be appended after \emph{main.py} 

\begin{verbatim}
python main.py --config config.dqn.cartpole
\end{verbatim}

When you type path of configuration file, you should omit '.py' in the name of the configuration file. If the command does not contain configuration path, it runs with the default configuration in the script. All parameters in the configuration file can be changed by using the parser without modifying the configuration file. To change the configuration using the command, users should type the parameter key and value. The example command for training PPO algorithm using sync distributed training with changing number of workers in distributed structure is as follows. 

\begin{verbatim}
python main.py --sync --config config.ppo.cartpole --train.num_workers 8
\end{verbatim}

With these simple commands, the algorithms and environments of JORLDY can be executed. 

\section{Conclusion}

This paper proposes a new open source reinforcement learning framework: JORLDY. It contains various RL algorithms and environments which can be fully customized. With this flexibility, users can not only modify the existing algorithms and environments but also add their algorithms and environments for their research. Also, JORLDY provides various types of distributed structure for training the RL algorithms. The users can select one of these distributed structures to proceed with training. 

We think that even non-experts on RL will be able to use RL algorithms more easily and apply them to their various environments by using JORLDY. We also believe that it will help researchers and students working on RL to advance their research. In order to contribute to the development of reinforcement learning, JORLDY will continue to develop in the future.

\section*{Acknowledgments}
We thank Kakao Enterprise AI Lab \& Service, Advanced tech team member for their support. Also, We appreciate Jinhyung Ahn who contributes the development of JORLDY a lot.  

%Bibliography
\bibliographystyle{unsrt}  
\bibliography{references}  

\clearpage

% Appendix
\appendix

\section{Example of configuration file}

% 여기에 내용적으로 추가할건 없을까요?

This is the example of configuration file. The following code is the configuration file of the DQN algorithm for CartPole environment. All the configurations of the algorithms for the environments can be found in the following link: \url{https://bit.ly/3tia1O0} 

\begin{lstlisting}
### DQN CartPole Config ###

env = {
    "name": "cartpole",
    "action_type": "discrete",
    "render": False,
}

agent = {
    "name": "dqn",
    "network": "discrete_q_network",
    "gamma": 0.99,
    "epsilon_init": 1.0,
    "epsilon_min": 0.01,
    "explore_ratio": 0.2,
    "buffer_size": 50000,
    "batch_size": 32,
    "start_train_step": 2000,
    "target_update_period": 500,
}

optim = {
    "name": "adam",
    "lr": 0.0001,
}

train = {
    "training": True,
    "load_path": None,
    "run_step": 100000,
    "print_period": 1000,
    "save_period": 10000,
    "eval_iteration": 10,
    # distributed setting
    "update_period": 32,
    "num_workers": 8,
}

\end{lstlisting}

\newpage

\section{Benchmark}

In the benchmark results, the bold scores are best performance among the implemented algorithms of JORLDY for each environment. All the scores are average value of the scores of three runs and each score is obtained using the final evaluation score with tensorboard smooth 0.6. All the metrics, specs and configurations can be found in the following link: \url{https://bit.ly/3tVYYJo} 

\subsection{JORLDY Benchmark on Atari}

All the agents for Atari environment are trained for 10M steps. For distributed training, 8 actors are used for PPO and 128 actors are used for Ape-X. 

\begin{table}[htbp]
\centering
\caption {Atari Benchmark of JORLDY}
\begin{adjustbox}{width=\textwidth}
\begin{tabular}{|c|c|c|c|c|c|c|c|c|c|c|c|c|}
\hline
& DQN & Double & Dueling & PER & NoisyNet & C51 & Multistep & RainbowDQN & QR-DQN & M-DQN & PPO & Ape-X \\ \hline
alien & 
\begin{tabular}[c]{@{}c@{}}1828\end{tabular} & 
\begin{tabular}[c]{@{}c@{}}626\end{tabular} & 
\begin{tabular}[c]{@{}c@{}}1791\end{tabular} & 
\begin{tabular}[c]{@{}c@{}}1505\end{tabular} & 
\begin{tabular}[c]{@{}c@{}}2192\end{tabular} & 
\begin{tabular}[c]{@{}c@{}}1445\end{tabular} & 
\begin{tabular}[c]{@{}c@{}}1510\end{tabular} & 
\begin{tabular}[c]{@{}c@{}}1997\end{tabular} & 
\begin{tabular}[c]{@{}c@{}}1930\end{tabular} & 
\begin{tabular}[c]{@{}c@{}}2026\end{tabular} & 
\begin{tabular}[c]{@{}c@{}}5779\end{tabular} &
\begin{tabular}[c]{@{}c@{}}\textbf{8437}\end{tabular} \\ \hline
assault & 
\begin{tabular}[c]{@{}c@{}}1841\end{tabular} & 
\begin{tabular}[c]{@{}c@{}}1421\end{tabular} & 
\begin{tabular}[c]{@{}c@{}}1799\end{tabular} & 
\begin{tabular}[c]{@{}c@{}}6241\end{tabular} & 
\begin{tabular}[c]{@{}c@{}}4399\end{tabular} & 
\begin{tabular}[c]{@{}c@{}}2064\end{tabular} & 
\begin{tabular}[c]{@{}c@{}}2496\end{tabular} & 
\begin{tabular}[c]{@{}c@{}}4898\end{tabular} & 
\begin{tabular}[c]{@{}c@{}}9088\end{tabular} & 
\begin{tabular}[c]{@{}c@{}}1267\end{tabular} & 
\begin{tabular}[c]{@{}c@{}}13967\end{tabular} &
\begin{tabular}[c]{@{}c@{}}\textbf{35022}\end{tabular} \\ \hline
battle zone & 
\begin{tabular}[c]{@{}c@{}}10297\end{tabular} & 
\begin{tabular}[c]{@{}c@{}}17983\end{tabular} & 
\begin{tabular}[c]{@{}c@{}}6139\end{tabular} & 
\begin{tabular}[c]{@{}c@{}}26724\end{tabular} & 
\begin{tabular}[c]{@{}c@{}}25858\end{tabular} & 
\begin{tabular}[c]{@{}c@{}}9540\end{tabular} & 
\begin{tabular}[c]{@{}c@{}}21183\end{tabular} & 
\begin{tabular}[c]{@{}c@{}}34364\end{tabular} & 
\begin{tabular}[c]{@{}c@{}}34416\end{tabular} & 
\begin{tabular}[c]{@{}c@{}}31044\end{tabular} & 
\begin{tabular}[c]{@{}c@{}}55152\end{tabular} &
\begin{tabular}[c]{@{}c@{}}\textbf{59721}\end{tabular} \\ \hline
breakout & 
\begin{tabular}[c]{@{}c@{}}295\end{tabular} & 
\begin{tabular}[c]{@{}c@{}}301\end{tabular} & 
\begin{tabular}[c]{@{}c@{}}322\end{tabular} & 
\begin{tabular}[c]{@{}c@{}}359\end{tabular} & 
\begin{tabular}[c]{@{}c@{}}364\end{tabular} & 
\begin{tabular}[c]{@{}c@{}}238\end{tabular} & 
\begin{tabular}[c]{@{}c@{}}58\end{tabular} & 
\begin{tabular}[c]{@{}c@{}}342\end{tabular} & 
\begin{tabular}[c]{@{}c@{}}339\end{tabular} & 
\begin{tabular}[c]{@{}c@{}}349\end{tabular} & 
\begin{tabular}[c]{@{}c@{}}388\end{tabular} &
\begin{tabular}[c]{@{}c@{}}\textbf{453}\end{tabular} \\ \hline
enduro & 
\begin{tabular}[c]{@{}c@{}}1060\end{tabular} &
\begin{tabular}[c]{@{}c@{}}1007\end{tabular} &
\begin{tabular}[c]{@{}c@{}}1001\end{tabular} &
\begin{tabular}[c]{@{}c@{}}1629\end{tabular} &
\begin{tabular}[c]{@{}c@{}}1430\end{tabular} &
\begin{tabular}[c]{@{}c@{}}1054\end{tabular} &
\begin{tabular}[c]{@{}c@{}}46\end{tabular} & 
\begin{tabular}[c]{@{}c@{}}3303\end{tabular} &
\begin{tabular}[c]{@{}c@{}}2713\end{tabular} &
\begin{tabular}[c]{@{}c@{}}1011\end{tabular} &
\begin{tabular}[c]{@{}c@{}}3621\end{tabular} &
\begin{tabular}[c]{@{}c@{}}\textbf{13340}\end{tabular} \\ \hline
ms pacman & 
\begin{tabular}[c]{@{}c@{}}2940\end{tabular} &
\begin{tabular}[c]{@{}c@{}}2024\end{tabular} &
\begin{tabular}[c]{@{}c@{}}2129\end{tabular} &
\begin{tabular}[c]{@{}c@{}}2205\end{tabular} &
\begin{tabular}[c]{@{}c@{}}2538\end{tabular} &
\begin{tabular}[c]{@{}c@{}}2408\end{tabular} &
\begin{tabular}[c]{@{}c@{}}1964\end{tabular} & 
\begin{tabular}[c]{@{}c@{}}2255\end{tabular} &
\begin{tabular}[c]{@{}c@{}}2639\end{tabular} &
\begin{tabular}[c]{@{}c@{}}2421\end{tabular} &
\begin{tabular}[c]{@{}c@{}}\textbf{7022}\end{tabular} &
\begin{tabular}[c]{@{}c@{}}6611\end{tabular} \\ \hline
pong & 
\begin{tabular}[c]{@{}c@{}}21\end{tabular} & 
\begin{tabular}[c]{@{}c@{}}21\end{tabular} & 
\begin{tabular}[c]{@{}c@{}}21\end{tabular} & 
\begin{tabular}[c]{@{}c@{}}21\end{tabular} & 
\begin{tabular}[c]{@{}c@{}}21\end{tabular} & 
\begin{tabular}[c]{@{}c@{}}21\end{tabular} & 
\begin{tabular}[c]{@{}c@{}}21\end{tabular} & 
\begin{tabular}[c]{@{}c@{}}21\end{tabular} & 
\begin{tabular}[c]{@{}c@{}}21\end{tabular} & 
\begin{tabular}[c]{@{}c@{}}21\end{tabular} & 
\begin{tabular}[c]{@{}c@{}}21\end{tabular} &
\begin{tabular}[c]{@{}c@{}}21\end{tabular} \\ \hline
qbert & 
\begin{tabular}[c]{@{}c@{}}13053\end{tabular} & 
\begin{tabular}[c]{@{}c@{}}17220\end{tabular} & 
\begin{tabular}[c]{@{}c@{}}15249\end{tabular} & 
\begin{tabular}[c]{@{}c@{}}11861\end{tabular} & 
\begin{tabular}[c]{@{}c@{}}16246\end{tabular} & 
\begin{tabular}[c]{@{}c@{}}15006\end{tabular} & 
\begin{tabular}[c]{@{}c@{}}14672\end{tabular} & 
\begin{tabular}[c]{@{}c@{}}27760\end{tabular} & 
\begin{tabular}[c]{@{}c@{}}17560\end{tabular} & 
\begin{tabular}[c]{@{}c@{}}14867\end{tabular} & 
\begin{tabular}[c]{@{}c@{}}\textbf{31684}\end{tabular} &
\begin{tabular}[c]{@{}c@{}}26768\end{tabular} \\ \hline
robotank & 
\begin{tabular}[c]{@{}c@{}}7\end{tabular} & 
\begin{tabular}[c]{@{}c@{}}15\end{tabular} & 
\begin{tabular}[c]{@{}c@{}}22\end{tabular} & 
\begin{tabular}[c]{@{}c@{}}25\end{tabular} & 
\begin{tabular}[c]{@{}c@{}}32\end{tabular} & 
\begin{tabular}[c]{@{}c@{}}27\end{tabular} & 
\begin{tabular}[c]{@{}c@{}}26\end{tabular} & 
\begin{tabular}[c]{@{}c@{}}37\end{tabular} & 
\begin{tabular}[c]{@{}c@{}}32\end{tabular} & 
\begin{tabular}[c]{@{}c@{}}24\end{tabular} & 
\begin{tabular}[c]{@{}c@{}}33\end{tabular} &
\begin{tabular}[c]{@{}c@{}}\textbf{63}\end{tabular} \\ \hline
seaquest & 
\begin{tabular}[c]{@{}c@{}}159\end{tabular} & 
\begin{tabular}[c]{@{}c@{}}124\end{tabular} & 
\begin{tabular}[c]{@{}c@{}}134\end{tabular} &
\begin{tabular}[c]{@{}c@{}}6000\end{tabular} &
\begin{tabular}[c]{@{}c@{}}120\end{tabular} &
\begin{tabular}[c]{@{}c@{}}3960\end{tabular} &
\begin{tabular}[c]{@{}c@{}}3622\end{tabular} & 
\begin{tabular}[c]{@{}c@{}}13377\end{tabular} &
\begin{tabular}[c]{@{}c@{}}20710\end{tabular} &
\begin{tabular}[c]{@{}c@{}}1974\end{tabular} &
\begin{tabular}[c]{@{}c@{}}2667\end{tabular} &
\begin{tabular}[c]{@{}c@{}}\textbf{62357}\end{tabular} \\ \hline
spaceinvaders & 
\begin{tabular}[c]{@{}c@{}}649\end{tabular} &
\begin{tabular}[c]{@{}c@{}}702\end{tabular} &
\begin{tabular}[c]{@{}c@{}}1302\end{tabular} &
\begin{tabular}[c]{@{}c@{}}995\end{tabular} &
\begin{tabular}[c]{@{}c@{}}872\end{tabular} &
\begin{tabular}[c]{@{}c@{}}1265\end{tabular} &
\begin{tabular}[c]{@{}c@{}}481\end{tabular} & 
\begin{tabular}[c]{@{}c@{}}1346\end{tabular} &
\begin{tabular}[c]{@{}c@{}}1160\end{tabular} &
\begin{tabular}[c]{@{}c@{}}901\end{tabular} &
\begin{tabular}[c]{@{}c@{}}2862\end{tabular} &
\begin{tabular}[c]{@{}c@{}}\textbf{102980}\end{tabular} \\ \hline
time pilot & 
\begin{tabular}[c]{@{}c@{}}2981\end{tabular} &
\begin{tabular}[c]{@{}c@{}}1729\end{tabular} &
\begin{tabular}[c]{@{}c@{}}3269\end{tabular} &
\begin{tabular}[c]{@{}c@{}}4649\end{tabular} &
\begin{tabular}[c]{@{}c@{}}3947\end{tabular} &
\begin{tabular}[c]{@{}c@{}}3904\end{tabular} &
\begin{tabular}[c]{@{}c@{}}4148\end{tabular} & 
\begin{tabular}[c]{@{}c@{}}9125\end{tabular} &
\begin{tabular}[c]{@{}c@{}}4545\end{tabular} &
\begin{tabular}[c]{@{}c@{}}3621\end{tabular} &
\begin{tabular}[c]{@{}c@{}}\textbf{16615}\end{tabular} &
\begin{tabular}[c]{@{}c@{}}13776\end{tabular} \\ \hline
\end{tabular}
\end{adjustbox}
\end{table}

\subsection{JORLDY Benchmark on Mujoco}

All the agents for MuJoCo environment are trained for 1M steps. 

\begin{table}[htbp]
\centering
\caption {Mujoco Benchmark of JORLDY}
\begin{adjustbox}{width=0.4\textwidth}
\begin{tabular}{|c|c|c|c|}
\hline
& DDPG & TD3 & SAC \\ \hline
Ant & 
\begin{tabular}[c]{@{}c@{}}673\end{tabular} & 
\begin{tabular}[c]{@{}c@{}}\textbf{5945}\end{tabular} &
\begin{tabular}[c]{@{}c@{}}4763\end{tabular} \\ \hline
HalfCheetah & 
\begin{tabular}[c]{@{}c@{}}4335\end{tabular} & 
\begin{tabular}[c]{@{}c@{}}\textbf{9305}\end{tabular} &
\begin{tabular}[c]{@{}c@{}}8385\end{tabular} \\ \hline
Hopper & 
\begin{tabular}[c]{@{}c@{}}1938\end{tabular} & 
\begin{tabular}[c]{@{}c@{}}3622\end{tabular} &
\begin{tabular}[c]{@{}c@{}}\textbf{3627}\end{tabular} \\ \hline
\begin{tabular}[c]{@{}c@{}}Inverted Double\\ Pendulum \end{tabular} & \begin{tabular}[c]{@{}c@{}}9325\end{tabular} & 
\begin{tabular}[c]{@{}c@{}}\textbf{9359}\end{tabular} &
\begin{tabular}[c]{@{}c@{}}9328\end{tabular} \\ \hline
Reacher & 
\begin{tabular}[c]{@{}c@{}}\textbf{-5.8}\end{tabular} &
\begin{tabular}[c]{@{}c@{}}-6\end{tabular} & 
\begin{tabular}[c]{@{}c@{}}-8.3\end{tabular} \\ \hline
Walker & 
\begin{tabular}[c]{@{}c@{}}3957\end{tabular} & 
\begin{tabular}[c]{@{}c@{}}\textbf{5129}\end{tabular} &
\begin{tabular}[c]{@{}c@{}}3987\end{tabular} \\ \hline
\end{tabular}
\end{adjustbox}
\end{table}
\end{document}